\documentclass[letterpaper, 10pt, journal, twocolumn, twoside]{IEEEtran}

\IEEEoverridecommandlockouts  

\usepackage{amsmath} 
\usepackage{amssymb}  

\usepackage{graphicx}
\usepackage{hyperref}
\usepackage{booktabs} 
\usepackage[T1]{fontenc}
\usepackage{newtxtext}  
\usepackage{pifont}
\usepackage{tikz}

\makeatletter
\def\ps@IEEEtitlepagestyle{%
  \def\@oddfoot{\mycopyrightnotice}%
  \def\@oddhead{\hbox{}\@IEEEheaderstyle\leftmark\hfil\thepage}\relax
  \def\@evenhead{\@IEEEheaderstyle\thepage\hfil\leftmark\hbox{}}\relax
  \def\@evenfoot{}%
}
\def\mycopyrightnotice{%
  \begin{minipage}{\textwidth}
  \centering \scriptsize
  Copyright~\copyright~2025 IEEE. Personal use of this material is permitted. Permission from IEEE must be obtained for all other uses, in any current or future media, including\\reprinting/republishing this material for advertising or promotional purposes, creating new collective works, for resale or redistribution to servers or lists, or reuse of any copyrighted component of this work in other works.
  \end{minipage}
}
\makeatother

\newcommand{\filledstar}{\ding{72}}
\newcommand{\redfilledstar}{%
\mbox{\hspace{0.3em}\protect\tikz[baseline=-0ex]\protect\fill[red] (0,0) rectangle (0.2cm,0.2cm);
\hspace{-1.505em}\filledstar\hspace{-0.1em}\filledstar
}}
\newcommand{\greenfilledstar}{%
\mbox{\hspace{0.3em}\protect\tikz[baseline=-0ex]\protect\fill[red] (0,0) rectangle (0.2cm,0.2cm);
\hspace{-1.52em}\filledstar\hspace{-0.1em}\filledstar\hspace{-0.4em}
}}
\usepackage[final]{changes}
\definechangesauthor[name={Karla}, color=teal]{KS}
\definechangesauthor[name={Jan}, color=orange]{JB}
\definechangesauthor[name={Petr}, color=brown]{PV}

\title{\LARGE \bf ILeSiA: Interactive Learning of Robot Situational Awareness from Camera Input}

\author{Petr Vanc$^{1,*}$, Giovanni Franzese$^{2}$, Jan Kristof Behrens$^{1}$, Cosimo Della Santina$^{2}$,\\Karla Stepanova$^{1}$, Jens Kober$^{2}$ and Robert Babuska$^{1,2}$
\thanks{Manuscript received: June 2, 2025; Revised:
August 12, 2025; Accepted: August 18, 2025. This paper was recommended for publication by Editor Markus Vincze upon evaluation of the Associate Editor and Reviewers’
comments.}
\thanks{*~The author conducted this research during his internship visit at TU Delft.}
\thanks{$^{1}$ Authors are with Czech Institute of Informatics, Robotics, and Cybernetics, Czech Technical University in Prague, Czech Republic {\tt\small \{petr.vanc, jan.kristof.behrens, karla.stepanova\}@cvut.cz}}%
\thanks{$^{2}$ Authors are with Cognitive Robotics, Delft University of Technology, The Netherlands {\tt\small \{G.Franzese, C.DellaSantina, J.Kober, R.Babuska\}@tudelft.nl}}.%
\thanks{This work was supported by the European Union under the
project Robotics and advanced industrial production (reg. no.
CZ.02.01.01/00/22\_008/0004590) and by the Czech Science Foundation (GACR), grant no. 21-31000S. P.V. was additionally supported by CTU Student Grant Agency (reg. no.
SGS23/179/OHK3/3T/13)}
\thanks{Digital Object Identifier (DOI): 10.1109/LRA.2025.3601037}}

\begin{document}

\maketitle

\begin{abstract}
Learning from demonstration is a promising approach for teaching robots new skills. However, a central challenge in the execution of acquired skills is the ability to recognize faults and prevent failures. This is essential because demonstrations typically cover only a limited set of scenarios and often only the successful ones. During task execution, unforeseen situations may arise, such as changes in the robot's environment or interaction with human operators. To recognize such situations, this paper focuses on teaching the robot situational awareness by using a camera input and labeling frames as safe or risky. We train a Gaussian Process (GP) regression model fed by a low-dimensional latent space representation of the input images. 
The model outputs a continuous risk score ranging from zero to one, quantifying the degree of risk at each timestep.
This allows for pausing task execution in unsafe situations and directly adding new training data, labeled by the human user. 
Our experiments on a robotic manipulator show that the proposed method can reliably detect both known and novel faults using only a single example for each new fault. In contrast, a standard multi-layer perceptron (MLP) performs well only on faults it has encountered during training.
Our method enables the next generation of cobots to be rapidly deployed with easy-to-set-up, vision-based risk assessment, proactively safeguarding humans and detecting misaligned parts or missing objects before failures occur.
We provide all the code and data required to reproduce our experiments at \href{http://imitrob.ciirc.cvut.cz/publications/ilesia}{imitrob.ciirc.cvut.cz/publications/ilesia}.
\end{abstract}

\section{Introduction}

Learning from demonstration has great potential to reduce the setup cost and increase the adaptability of robotic systems. This typically involves guiding a robot kinesthetically through a task to provide a nominal demonstration and teach the robot a \textit{skill} that it is then expected to reproduce, for example, picking up a peg. However, every recorded demonstration implicitly assumes specific task conditions, for example, that the peg is in a known location. During execution, these hidden assumptions may no longer hold, raising the critical question: at each moment, how can the robot tell whether it is safe to continue?

\vspace{-0.9em}
A major challenge is distinguishing harmless variations from those that could lead to a task \textit{failure}: an inability of the robot to complete the intended action successfully. For instance, a slight rotation of the peg might still allow for successful grasping, whereas a larger misalignment might make the peg unpickable. 
Identifying such potentially harmful deviations requires task-dependent situational awareness far beyond simply following the recorded trajectory.

To allow the robot to gain situational awareness, we assume that there is a human supervisor capable of detecting potential faults.
The term \textit{fault} \cite{loborg1994error} refers to 
any condition that can lead to a task failure, such as a peg not present in the expected location. By detecting such conditions, we may prevent pending failures, such as the peg not being grasped.
Faults are treated as root causes; once detected, they prompt manual recovery without requiring the system to diagnose their origin.

We aim to detect both \textit{seen faults} (those encountered and labeled during training) and \textit{novel faults} (out-of-distribution scenarios not previously observed). The human supervisor may interactively provide feedback to the autonomously acting robot by detecting faults.

\begin{figure}
\centering
\includegraphics[width=0.485\textwidth]{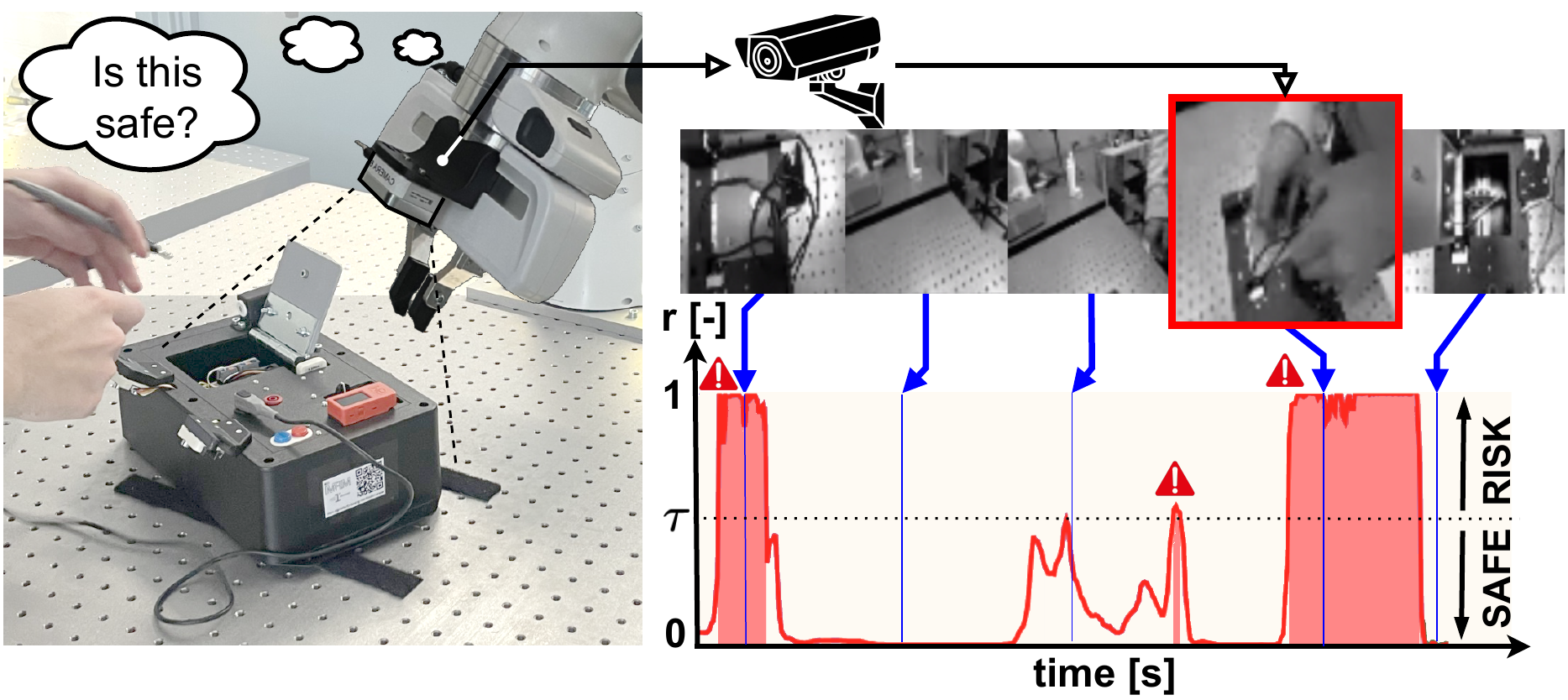}
\vspace{-2.1em}
\caption{An illustration showing the use of the ILeSiA system. The plot shows an estimated \textit{risk score} ($r$) during the skill execution. The risk score in each timestep is estimated from the corresponding camera image. If the threshold $\tau$ is exceeded, \textit{risk flag} is raised and the execution is paused. In the illustration, a camera mounted on the robot sees the human hand and detects the situation as \textit{risky}. Manipulation tasks use the Electronic Task Board from Robothon 2023 to demonstrate various skills.}
\label{fig_robothonbox}
    \vspace{-1.1em}
\end{figure}

Unlike recent situation-awareness methods that rely on large, previously collected datasets with extensive annotations \cite{Ruiz-Celada_Dalmases_Zaplana_Rosell_2024, DERNER2021103676}, our interactive framework labels faults during task execution. This human-in-the-loop strategy enables the system to detect both familiar and unexpected faults with only a handful of examples.

To demonstrate the effectiveness of our framework, we validate our approach on the Electronic Task Board from Robothon 2023\footnote{\label{lfdmodule}Platonics: Robothon 2023 website: 
\href{https://platonics-delft.github.io/}{platonics-delft.github.io/}
} (see Fig.~\ref{fig_robothonbox}, left). 
The onboard camera continuously feeds visual data to our risk estimator, which outputs a scalar \textit{risk score} $r$ quantifying deviations from safe states. 
When this score exceeds a predefined threshold, a potential fault is indicated, such as ``door not open'', and the system pauses the execution to prevent a downstream failure, such as an attempt to manipulate an object behind the unopened door. 

When the robot encounters an out-of-distribution visual input (i.e., a {novel fault}), it pauses the execution and requests the human supervisor to label the situation as either \textit{safe} or \textit{risky}. These sparse annotations provided by the supervisor are employed to update our risk estimator’s model, improving its predictions in subsequent executions. This approach enables effective online ground-truth labeling without the need to manually annotate thousands of frames in advance.

We call this approach ILeSiA: Interactive Learning of Situational Awareness from Camera Input. 
ILeSiA starts with a single fault free demonstration captured by a camera and then enters an interactive cycle of autonomous execution and sparse human feedback. Whenever the robot encounters a novel, out-of-distribution condition, human supervisor adds new annotations to fine-tune the risk estimator’s model. Over successive runs, the system incrementally builds internal knowledge and becomes skilled at recognizing a variety of typical faults in its specific environment with ever fewer interventions. 

Because the model is trained only on task-relevant examples, it remains lightweight and efficient, enabling reliable real-time risk detection suitable for industrial applications and human–robot collaboration. By allowing the supervisor to define what constitutes a risky situation, ILeSiA enables a flexible, supervisor-guided expansion of the robot’s situational awareness over time.

The main contributions of this paper are:
\begin{itemize}
    \item Development of a compact risk estimator method: it estimates the risk score $r$ (evidence of {fault}) at any given timestep based on visual input from the camera, requiring only a single demonstration of the {fault}. It continuously assesses the environment for potential faults that could hinder the robot's task execution.
    \item Integration of the risk estimator into a Learning from Demonstration (LfD) Framework for interactive aggregation of new {faults} on top of typical robotic skills executions, automatic model retraining, and establishing a human-in-the-loop system, where a human supervisor annotates new and previously unseen faults as either safe or risky, and expanding the robot's knowledge.
\end{itemize}

All source code, interactive visualization tools, experimental videos, and datasets are available at: 
\href{http://imitrob.ciirc.cvut.cz/publications/ilesia}{\texttt{imitrob.ciirc.cvut.cz/publications/ilesia}}.

\vspace{-0.8em}
\section{Related Work}
\label{sec:sota}

The field of fault detection in robotics has traditionally focused on preventing mechanical or operational failures through real-time processing. Early works by Van and Ge \cite{Van_Ge_Ren_2017} and Wu et al. \cite{Wu_Luo_Zeng_Li_Zheng_2016} demonstrated the value of using sensor data to detect deviations from normal behavior \cite{chandola2009anomaly}. Park et al. \cite{7487160} and Lello et al. \cite{6697200} trained hidden Markov models on multi-modal observations, e.g., force or proprioception, to detect departures from a nominal manipulation trajectory. Other data-driven approaches employ one-class classifiers such as linear SVMs to flag failures without requiring their examples during training \cite{kappler2015data}.

More recent efforts have broadened the scope from purely operational failures to encompass situational faults inferred from visual input, such as detecting a human hand entering the robot’s workspace. Ruiz-Celada et al. \cite{Ruiz-Celada_Dalmases_Zaplana_Rosell_2024} integrated perception and reasoning to capture epistemic uncertainty and enable robots to recognize when their sensory information or model knowledge is insufficient \cite{hullermeier2021aleatoric}.

A closely related challenge is anomaly detection (AD) viewed as an out-of-distribution (OOD) problem: learning a model of ``normal'' data (often with only positive examples) and scoring deviations \cite{8279425}, or detecting anomalies semantically by using LLMs \cite{sinhaAnomaly2024}.

In contrast, our risk-estimation framework combines explicitly labeled risky examples for seen faults with OOD cues. Instead of relying on an $N+1$ classification scheme that treats risk as a separate class or assuming that risky events are vanishingly rare, we continuously evaluate the incoming camera stream, using both learned fault patterns and epistemic uncertainty.

Vision-based failure detection methods have also been explored independently of AD. Inceoglu et al. \cite{8594169} encoded visual features by using histograms of oriented gradients and applied PCA to detect proprioceptive failures from camera images, while others compare outputs of different convolutional neural networks to identify faults in real time \cite{10124202}.

Further bridging the gap between traditional fault detection and modern risk assessment, interactive imitation learning (IIL) techniques \cite{9076630, 9636710} provide a framework for robots to learn complex tasks through human feedback \cite{iilsurvey}. 
Techniques such as DAgger \cite{daggeroriginal}, ThriftyDAgger \cite{ThriftyDAgger}, and follow-up work \cite{8793698, 8968287} optimize the learning process by focusing on scenarios where human intervention is most critical, thereby reducing the expert burden, as explored in FIRE (Failure Identification to Reduce Expert Burden) \cite{FIRE}.

Unlike other approaches, our approach automatically learns when a situation is risky or safe. For example, our model can be trained to recognize that an open door is risky at the beginning of execution. 

Recent one-class, uncertainty-aware detectors like FAIL-Detect \cite{xu2025detectfailuresfailuredata} introduce a two-stage pipeline: first distilling image features and robot states into scalar uncertainty scores, then applying time-varying conformal prediction bands to flag faults without needing data for these faults. Similarly, the TAMPURA framework \cite{curtis2024partiallyobservabletaskmotion} learns probabilistic transition models over abstract belief states via simulation and uses Bayesian upper confidence bounds to direct exploration toward safer, more informative plans.

Overall, while traditional studies lay a solid foundation for understanding and detecting faults in robotic systems, our work extends these concepts by detecting both learned and out-of-distribution faults via real-time video analysis and interactive, task-relevant fault labeling. This enables the system to alert a human supervisor to situations that would otherwise be deemed unsafe and to simultaneously collect new training samples.

\section{Method}
\label{sec:method}

\begin{figure}[t]
\centering
\includegraphics[width=0.49\textwidth]{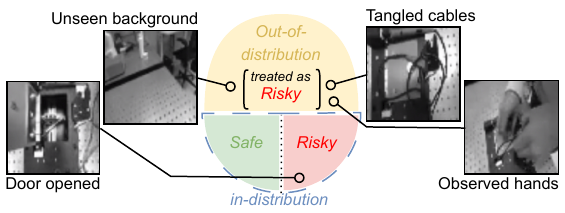}
\vspace{-2.0em}
\caption{Venn diagram example of output space. Images are classified as {in-distribution} (i.e., similar to the seen images) or {out-of-distribution}. Our model classifies images as \textit{safe} or potentially \textit{risky}, where potentially risky cases are presented to the human supervisor to specify them as risky or safe and update its own knowledge about risk estimation.}
\label{fig_varspace}
\end{figure}

The goal of our method is to accurately classify each incoming image observed during skill execution as safe or risky. However, the underlying categorization includes three classes: safe, risky, and out-of-distribution (see Fig.~\ref{fig_varspace}). The safe and risky categories are in-distribution, meaning that similar images were covered by the training data. In contrast, out-of-distribution situations (i.e., images that differ significantly from anything previously seen) are treated as risky by default.

We propose a two-stage approach. In the first stage, each image is encoded into a low-dimensional feature representation. In the second stage, a risk score is estimated from these features. This design is motivated by our aim to capture epistemic uncertainty in the current state: uncertainty arising from a lack of knowledge. In this way, out-of-distribution inputs are treated as novel (potentially risky) situations.

In the first stage, we adopt a standard autoencoder (AE) \cite{HintonSalakhutdinov2006b} to extract compact and informative features from images. 
While standard autoencoders do not explicitly model uncertainty, they can serve as effective embeddings of general features into a latent space that may help classify out-of-distribution cases.
Training the autoencoder on all saved safe instances and \textit{seen faults} enhances the model's ability to recognize subtle visual variations during skill execution, making the learned feature space more sensitive to novel faults. 

In the second stage, we use Gaussian Process (GP) \cite{Rasmussen2004, Deisenroth2015} regression to estimate a risk score from the AE-encoded features. GPs provide both a fault prediction and a principled estimate of epistemic uncertainty thanks to their strong interpolation capabilities, making them well-suited for identifying diverse types of faults.

Note that replacing this two-stage approach with a variational autoencoder (VAE) would enable modeling of aleatoric uncertainty -- that is, uncertainty arising from inherent noise or variability in the data -- rather than epistemic uncertainty, which is the focus of our approach.

\subsection{Policy Definition} %
\label{sec:kin_demo}
We assume that a trajectory for an individual skill is recorded by kinesthetic teaching using the existing Learning from Demonstration (LfD) module\textsuperscript{\ref{lfdmodule}}, originally developed for the \emph{2023 Robothon challenge}, see Fig.~\ref{fig_robothonbox}. The camera mounted on the end effector is recording a video during this demonstration. The LfD module enables the teaching and execution of robotic skills through kinesthetic demonstrations. We execute the learned skills and recorded images at $20 \text{Hz}$ to ensure a comprehensive visual record.

These newly acquired skills are stored and later performed towards a specific object (object-centric) as fixed trajectories within the task space coordinates. This ensures consistent camera views during the trajectory execution, even when the target object moves between individual demonstrations.

\subsection{State Representation}

\begin{figure}[t]
\centering
\includegraphics[width=0.49\textwidth]{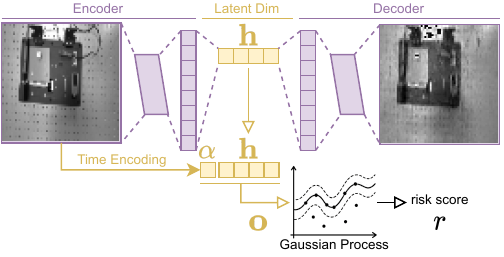}
\vspace{-2em}
\caption{Video embedding architecture utilizing a 4-layer autoencoder and its connection to the risk estimator.}
\label{fig_sim}
\end{figure}

To effectively detect faults during a novel demonstration, we first process the video signal, resize it, and convert it to grayscale. The videos were recorded in laboratory conditions, eliminating concerns about image lighting sensitivity or camera shake. Subsequently, we embed the input images in a latent space utilizing an autoencoder network (see Fig.~\ref{fig_sim}). 

\subsection{Risky or Safe: Learning and Judging The Current Situation}
\label{sec:risky_safe}

In the second stage, we estimate the risk score for the observed situation. We create an input vector $\mathbf{o}$ for the risk estimator by concatenating the latent space vector $\mathbf{h}$ corresponding to the frame from the demonstration at normalized time $\alpha$ (positional encoding): 
\begin{equation}
    \mathbf{o} = \mathbf{h} \oplus \alpha,
\end{equation}
where $\oplus$ denotes concatenation. The risk score $r$ for each skill is computed in real time by using the following formula:
\begin{equation} 
r = \mathcal{R}(\mathbf{o}), \;\;\; r \in [0,1], 
\end{equation}
where $\mathcal{R}$ denotes the method used for risk estimation. 
We propose the use of a GP \cite{Rasmussen2004, Deisenroth2015} as a core component of our risk estimation method. A GP is defined as:
\begin{equation}
    f(\mathbf{x}) \sim \mathcal{GP}(m(\mathbf{x}), k(\mathbf{x}, \mathbf{x}')),
\end{equation}
where $m(\mathbf{x})$ is the mean function specified as $m(\mathbf{x}) = 0$ and $k(\mathbf{x}, \mathbf{x}')$ is the covariance function specified as a Radial Basis Function (RBF) kernel with Automatic Relevance Determination (ARD) \cite{Williams05} as:
\begin{equation}
    k(\mathbf{x}, \mathbf{x}') = \sigma_p^2 \exp\left(-\frac{1}{2}\sum_{d=1}^{D} \left(\frac{(x_d - x_d')^2}{\lambda_d^2}\right)\right),    
\label{eq:ard}
\end{equation}
where $\sigma_p^2$ is the prior uncertainty of the model. $x_d$ denotes the $d$-th component of $\mathbf{x}$. The ARD optimizes the horizontal and vertical \emph{length scale} parameters ($\lambda_d$, $\sigma_p^2$) that determine how quickly the model moves out of distribution. The optimization decides which features (latent variables) are important for predicting the output variable and which can be safely ignored.

The central challenge lies in handling novel situations. Novel visual inputs fall outside the trained distribution and are identified by their deviation from the training data, leveraging the model's inherent uncertainty.

When making posterior predictions (function $\mathbf{f}^*$) at new points $\mathbf{x}^*$ based on training data $\mathbf{X}$ and their ground truth fault (risky or safe) labels $\mathbf{y}$, the GP provides a posterior predictive distribution, which is also multivariate normal:
\begin{equation}
    \mathbf{f}^* | \mathbf{X}, \mathbf{y}, \mathbf{x}^* \sim \mathcal{N}(\mathbf{\mu}^*, \Sigma^*)    
\end{equation}
where:
{\footnotesize
$\mathbf{\mu}^* = K(\mathbf{x}^*, \mathbf{X}) [K(\mathbf{X}, \mathbf{X}) + \sigma_n^2 I]^{-1} \mathbf{y}$
and
$\sigma^{*2} = K(\mathbf{x}^*, \mathbf{x}^*) - K(\mathbf{x}^*, \mathbf{X}) [K(\mathbf{X}, \mathbf{X}) + \sigma_n^2 I]^{-1} K(\mathbf{X}, \mathbf{x}^*)$},
with $\sigma_n^2$ the observation-noise variance parameter computed from all training samples (the Gaussian likelihood's noise variance). 
$K(\mathbf{X},\mathbf{X})$ is the covariance matrix computed from the RBF kernel over all pairs of inputs in $\mathbf{X}$ and $I$ is an identity matrix. 

Following the human supervisor approach to handling novel situations, we classify anything that is not deemed safe as risky -- this includes both novel situations and scenarios corresponding to previously labeled faults. To capture this by our model, we propose a novel method for estimating the risk score $r$ of the given situation: 
\begin{equation}
\label{eq:novel_risk}
r = clip(\mu^* + \sigma^*),
\end{equation}
where $\mu^*$ (reflects the in-distribution behavior) and $\sigma^*$ (proxy for out-of-distribution detection) are parameters of the predicted posterior distribution provided by the GP, function $\mathrm{clip}(x) = \min(1, \max(0, x))$ clips the value between zero and one. Note that we use GP regression to obtain the distribution parameters $\mu$ and $\sigma$ from which we compute the real number $r$ \eqref{eq:novel_risk}. 

Finally, the binary variable indicating the presence or absence of {fault}, the \textit{risk flag} $y$, is computed given the selected threshold $\tau = 0.5$, chosen because the model is trained on binary labels and $\tau = 0.5$ represents the midpoint. Then,
\begin{equation}
y = \begin{cases} 
1 & \text{if } r > \tau \\ 
0 & \text{if } r \leq \tau 
\end{cases}.
\end{equation}

We didn't find any benefit in optimizing the $\tau$ or using separate threshold values for $\mu$ and $\sigma$, where the risk flag is raised if either $\mu$ or $\sigma$ exceeds its respective threshold, i.e., $y = (\mu^*>\tau_\mu) \vee (\sigma^*>\tau_\sigma)$.

\textit{Stop Signal:}\label{stop_signal} When the risk flag is detected (e.g., an unexpected object or human hand is observed), the execution is paused by issuing a stop signal that asks the human supervisor to confirm or reject the detected fault.
The execution continues after the fault is assessed by the human supervisor.

\subsection{Active and Interactive Labeling of Situations from Human Feedback}
\label{sec:labeling}

\begin{figure}[t]
\centering
\includegraphics[width=0.45\textwidth]{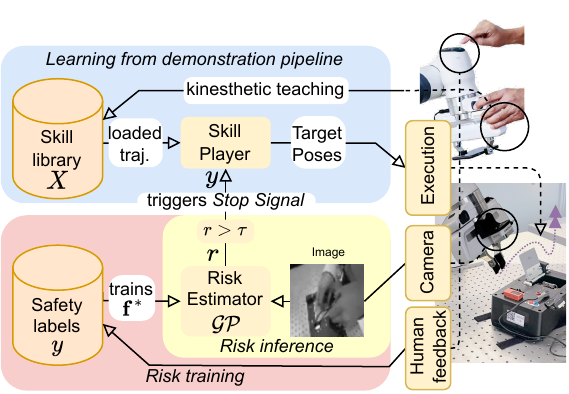}
\vspace{-0.6em}
\caption{ILeSiA: Interactive learning loop. 
 The proposed system allows the human supervisor to kinesthetically teach a new skill and execute it on the robot. If a fault arises during execution, the stop signal is triggered by the risk estimator or the human supervisor. This pauses the execution, and the human supervisor gives feedback. These safety labels are linked to the corresponding camera images and used to train the risk estimator.}
\vspace{-1em}
\label{fig_ldfimplementation}
\end{figure}

The risk estimator and its integration into the existing Learning from Demonstration module (ILeSiA)\textsuperscript{\ref{lfdmodule}} are shown in Fig.~\ref{fig_ldfimplementation}. The human supervisor halts the execution of the task upon detecting a fault. The risk estimator automatically captures these moments and uses them as training samples to update its future predictions. 
Labeling is facilitated through interactive inputs, for example, from the keyboard or using the robot's integrated buttons\footnote{\label{franka_buttons_git}Franka Buttons package: \href{https://github.com/franzesegiovanni/franka_buttons}{github.com/franzesegiovanni/franka\_buttons}}. This method allows for immediate and accurate labeling of relevant data points as safe or risky during demonstrations.

\section{Experimental Setup}
\label{sec:experimental-setup}

We tested the proposed system on manipulation tasks using the Electronic Task Board, which was part of the Robothon Challenge 2023 \cite{so2024digital}, see Fig.~\ref{fig_robothonbox}. The setup features a Franka Robotics Panda robot with an Intel RealSense D455 camera mounted close to the end-effector. 


\subsection{Recording Manipulation Skills}
\label{skills}

\begin{figure}
    \centering
    \includegraphics[trim={0cm 0cm 0cm 0.0cm},clip,width=1.0\linewidth]{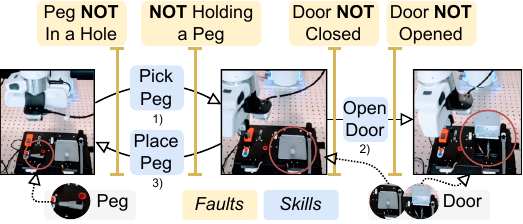}
    \vspace{-2em}
    \caption{Examples of recorded manipulation skills and their associated \textit{seen} faults, which define conditions for successful skill execution. When these faults occur, the skills result in failure. Videos demonstrating skill executions are available on our website\textsuperscript{\ref{fn:video}}.
    }
    \label{fig:preparedskillconditions}
    \vspace{-1em}
\end{figure}

The experiments focus on the three most frequently performed skills with the board: 1) \textit{Pick Peg}, 2) \textit{Open Door}, and 3) \textit{Place Peg}. The robot encounters the following situations (see Fig.~\ref{fig:preparedskillconditions}), which may be considered risky depending on the specific task:
\begin{enumerate}
\item Door open or closed.
\item Peg missing or misoriented during picking.
\item Peg misoriented after placement.
\item Cable held by the gripper after a picking attempt.
\item Obstacles or clutter, such as tangled cables.
\item Human hands visible in the camera frame.
\end{enumerate}

We categorize these situations into \emph{novel faults} (situations 5 and 6), i.e., faults for which the system has not been trained, and \emph{seen faults}, which are faults included and labeled within the training data.
However, even situations 1–4 contain a degree of novelty due to inherent variability: for example, the peg rotation can vary continuously, yet only a limited subset of rotations appears in the dataset.

\textit{Training Dataset}: For each task, we first recorded a kinesthetic demonstration. During trajectory recording, we ensured that the camera captured sufficient information for the human supervisor to identify faults from the video stream. We then recorded between 4 and 10 autonomous executions of each skill for training: at least one was safe, while the others were intentionally made to exhibit some of the faults (seen faults 1–4 listed above). Human supervisors provided labels for skill preconditions and execution faults.

\textit{Novel \& Seen Dataset}:\label{sec:datasets} 
For each task, we recorded 30 test executions simulating real-world scenarios containing seen faults, and between 8 and 11 executions containing novel faults (e.g., human hands or clutter entering the view). 
Each skill includes two evaluation segments to assess possible faults: one verifying preconditions at the beginning (e.g., verifying that the door is initially closed before opening) and another verifying postconditions at the end (e.g., confirming that the door is successfully opened).

In total, we recorded 3 kinesthetic demonstrations, 18 training executions (including various faults), 90 test executions (containing 190 instances of seen faults), and 30 novel test executions (containing 100 novel fault instances).
The datasets totaling 7.6 GB are publicly available on our website.\footnote{\label{fn:video}Website with datasets, skill execution videos, and interactive tool: \\
\href{http://imitrob.ciirc.cvut.cz/publications/ilesia/}{\texttt{http://imitrob.ciirc.cvut.cz/publications/ilesia/}}}
The observed inference time for detecting the risk score from an image varies around $35\pm 15$~ms. Training the GP model from the dataset \ref{sec:datasets} for a single skill took between $16$~s and $36$~s, depending on the number of input samples and the type of the skill: Place Peg $16$~s for (1670 samples), Open Door $33$~s (4050 samples), and Pick Peg $35$~s (2200 samples).

\subsection{Video Embedding} \label{video-embedding}

Our approach employs a standard autoencoder for dimensionality reduction, as detailed in Sec.~\ref{sec:method}. 
Its architecture comprises convolutional layers, normalization, ReLU activations, dropout, and max-pooling layers to compress the video data. We train the autoencoder on all skill executions except those containing novel faults (skills 5 and 6; see Sec.~\ref{skills}). This ensures that all \textit{seen faults} are accurately encoded.

\begin{table}[ht]
\centering
\caption{Comparison of video embedding approaches.}
\label{tab:embedding_comparison}
\begin{tabular}{lcc}
\toprule
& \textbf{Autoencoder} (ours) & \textbf{ResNet-50} \cite{Koonce2021} \\
\midrule
Input Size & $64 \times 64$ (grayscale) & $224 \times 224$ (RGB) \\
Preprocessing & Resize, grayscale & Upsample, normalize \\
Architecture & 4-layer CNN AE & Pre-trained CNN \\
Embedding & $\mathbb{R}^{12}$ (latent space size) & Blocks {2,3} ($\mathbb{R}^{{256,512}}$) \\
\bottomrule
\vspace{-1em}
\end{tabular}
\end{table}

\textit{ResNet Comparison}: We evaluated our autoencoder against a ResNet-50 model \cite{Koonce2021} pre-trained on ImageNet \cite{imagenet}. To match inputs, we upsampled and normalized the grayscale images to RGB using a per-channel transformation $x'_{c} = (x_{c} - \mu_{c})/\sigma_{c}$, with $\boldsymbol{\mu}=[0.485,0.456,0.406]$ and $\boldsymbol{\sigma}=[0.229,0.224,0.225]$, aligning the input distribution with the RGB ImageNet statistics expected by the pretrained ResNet. We extracted features from ResNet’s block 2 (256-dimensional) or block 3 (512-dimensional) and fed them into the same risk estimator. Table \ref{tab:embedding_comparison} summarizes the model specifications. 
On the \textit{seen dataset} (Sec.\ref{sec:datasets}) using the multi-layer perceptron (MLP) risk estimator (Sec.\ref{mlpdef}), ResNet-50 performed comparably to our autoencoder. Low-level features from earlier blocks excelled at detecting simpler faults (e.g., a held cable), whereas high-level features from later blocks were more effective at distinguishing semantic faults (e.g., open versus closed door).

We selected the autoencoder because it learns task-specific features unsupervised from our own data, dynamically adapting to the specific faults of interest. A more detailed embedding comparison is beyond the scope of this paper.

\subsection{Risk Estimator Dataset Collection}
\label{sec_risk_detector_dataset_collection}

The final dataset $\mathcal{D}$ used for training the risk estimator consists of $T$ recorded skill executions $t_j$ (see dataset details in Sec.~\ref{sec:datasets}): $$\mathcal{D} = \{\,t_j\}_{j=1}^{T},$$

Each skill execution $t_j$ comprises \(N_j\) frames:

$$t_j = \{\,d_{j,i}\}_{i=1}^{N_j},$$

where $d_{j,i}$ is i-th frame of the j-th execution trial:
$$
d_{j,i} = \bigl(\mathbf{h}_{j,i},\,R_{j,i},\,S_{j,i},\,\alpha_{j,i}\bigr),
$$
with $\mathbf{h}_{j,i}$ denoting the feature vector of frame $i$ in execution $j$, $R_{j,i} \in\{0,1\}$ indicating the “risky” label, $S_{j,i} \in\{0,1\}$ indicating the “safe” label, and $\alpha_{j,i} = \tfrac{i}{N_j} $ representing the normalized time within the execution.

All frames are considered \textit{safe} if no risk flag is triggered during the execution. Additionally, all frames recorded during the nominal kinesthetic demonstrations are labeled as \textit{safe}. To ensure that the GP estimator functions correctly, even with only a single trajectory, we include white and black image samples labeled as \textit{risky}; without these samples, the estimator would consistently predict "safe.". For executions containing a \textit{fault}, we prefer to use the $\mathcal{D}_{\text{selected}}$ dataset, which includes only explicitly labeled samples for training. This avoids introducing conflicting labels into the dataset, as we cannot precisely determine where a fault started or ended.
\begin{equation}
    \mathcal{D}_\text{selected} = \{ (d_i \in \mathcal{D} \mid (R_i \lor S_i) = 1 \}\text{.}
\end{equation}
\begin{table}[ht]
\centering
\caption{Comparison of GP and MLP performance on seen vs.\ novel faults.}
\label{tab:method_comparison}
\begin{tabular}{lccc}
\toprule
 & \textbf{Pick Peg} & \textbf{Open Door} & \textbf{Place Peg} \\
\midrule
\multicolumn{4}{c}{\textit{Seen-Fault Accuracy (\%)} (\(180\) samples total)} \\
\midrule
GP (ours)                & \textbf{91.8} & \textbf{99.3}          & \textbf{94.6}          \\
MLP                      & 90.5          & 91.5 & 93.9          \\
Logistic Regression (LR) & 71.2          & 69.3          & 79.7          \\
\midrule
Total seen-fault samples & 60 & 60 & 60 \\
\midrule
\multicolumn{4}{c}{\textit{False-Alarm NPV (\%)} (Negative predictive value)} \\
\midrule
GP (ours) & 91.8         & \textbf{95.7} & 99.9 \\
MLP       & \textbf{99.9}& 95.4          & \textbf{100.0}         \\
\midrule
\multicolumn{4}{c}{\textit{Novel-Fault Detection Accuracy (\%)} (\(100\) samples total)} \\
\midrule
GP (ours) & \textbf{84.2} & \textbf{92.3} & \textbf{73.9} \\
MLP       & 39.5           & 28.2          & 52.1           \\
\midrule
Total novel-fault samples & 38 & 39 & 23 \\
\bottomrule
\end{tabular}
\end{table}
\vspace{-15pt}

\section{Experiments}

In this section, we present experiments conducted to validate the proposed method and its key design decisions. The experiments evaluate three different risk-estimation methods using the datasets described in Sec.~\ref{sec:experimental-setup}. Sec.~\ref{sec:experiment_configuration} describes hyperparameter selection and training procedures. 


\subsection{Experiment Configuration}
\label{sec:experiment_configuration}

These are the baseline methods and their hyperparameters:

\subsubsection{Risk Estimation Baseline}\label{mlpdef} We compared our proposed GP-based method (see Sec.~\ref{sec:risky_safe}) with two baseline approaches: a MLP and logistic regression (LR). The MLP consists of three hidden layers, each containing 32 nodes with dropout applied. Both the MLP and LR were trained using binary cross-entropy loss optimized via Adam \cite{kingma2017adammethodstochasticoptimization} with a learning rate of $10^{-3}$ and weight decay of $10^{-5}$.

\subsubsection{Size of Latent Space} 

The size of the autoencoder's latent space, denoted as $\operatorname{len}(\mathbf{h})$, significantly influences the overall performance of our proposed method. We evaluate the embedding quality by assessing image reconstruction performance. The latent space must be sufficiently large to encode fault-capturing features; however, increasing its dimensionality adversely affects GP performance. These trade-offs are discussed in Sec.~\ref{sec:exp_reconstruction}. Based on initial experiments, we selected a latent space dimension of 12.

\subsubsection{Number of Training Epochs and Learning Rate} 

The number of training epochs may be dynamically determined based on validation loss computed from an independent demonstration trial. Typically, we train for more than 400 epochs with a learning rate ($lr$) of $0.01$. Note that the optimal number of epochs depends on the specific autoencoder architecture.

Importantly, training for fewer than 200 epochs at $lr = 0.01$ may lead to poor interpolation of the GP's length scale, hindering the effective detection of novel (out-of-distribution) faults.

\subsection{Performance on Seen Faults}
\label{sec_experiment_1}

\begin{figure}[t]
\centering
\hspace{1.5em}
\includegraphics[trim={0cm 0.0cm 0cm 0.0cm},clip,width=0.83\linewidth]{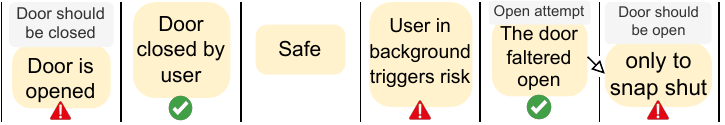}
\includegraphics[trim={1.5cm 0.9cm 0.9cm 1.8cm},clip,width=0.99\linewidth]{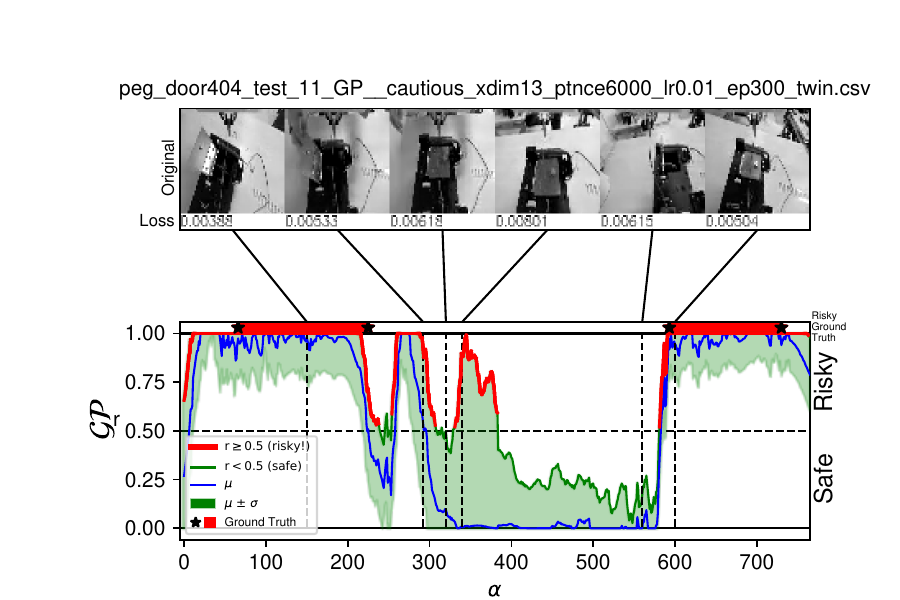}
\includegraphics[trim={1.5cm 0 0.9cm 5.45cm},clip,width=0.99\linewidth]{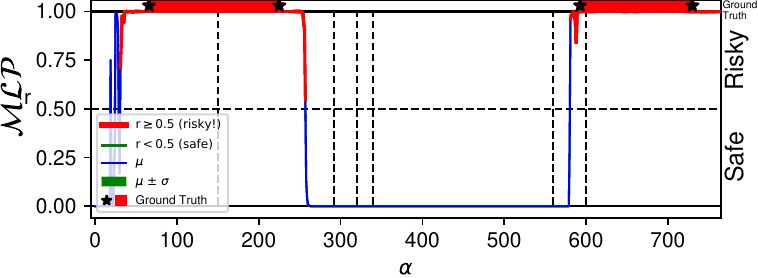}
\vspace{-1.5em}
\caption{Risk score prediction on a test trial for the \textit{Open Door} skill, comparing GP and MLP models. The red curve indicates predicted risk flags ($r > \tau$), while the blue curve represents safe predictions. Ground-truth risky segments are marked with \redfilledstar, emphasizing the importance of accurate detection during these intervals. Representative original frames and their corresponding reconstruction loss values are shown. For this skill, the door must start closed and end open to be considered safe. Midway through the trial, the human supervisor manually closes the door, correctly reducing the risk score. Later, the robot fails to open the door, and the risk score rises as the door snaps shut. 
The system employs a non-interrupting policy, allowing execution to continue despite triggered risk flags.}
\label{fig_modelcomparison1}
\vspace{-10pt}
\end{figure}

First, we assessed the model’s ability to detect seen faults, i.e., faults present in the training data, within held-out test trajectories. We compared GP and MLP risk estimators by evaluating their frame-by-frame predictions during two distinct fault events in a single test demonstration (Fig. \ref{fig_modelcomparison1}). Both models were trained using samples whose timestamps approximately aligned with our ground-truth labels. For example, since the door was already open before the human supervisor marked it as risky, it was reasonable for the model to label that region as high-risk, even beyond the explicitly labeled area.

Next, using the dataset from Sec. \ref{skills} (Table \ref{tab:method_comparison}), the GP model consistently achieved high accuracy on the held-out test demonstrations. For the \textit{Open Door} skill, which failed in 33\% of executions, the model correctly predicted 99\% of cases. All faults were detected, with only a single false alarm. Most remaining misclassifications occurred when test executions deviated from the training distribution, occasionally triggering false positives. Importantly, augmenting the training set with a few of these out-of-distribution executions improved the overall accuracy to nearly 100\%, except for a small number of cases affected by poor video reconstruction. Such degraded reconstructions (see Sec. \ref{sec:exp_reconstruction}) can be automatically flagged by comparing each frame’s reconstruction loss to the distribution observed during training.

\begin{figure}
    \centering
    \includegraphics[trim={0.4cm 0.0cm 0.4cm 0.0cm},clip,width=0.99\linewidth]{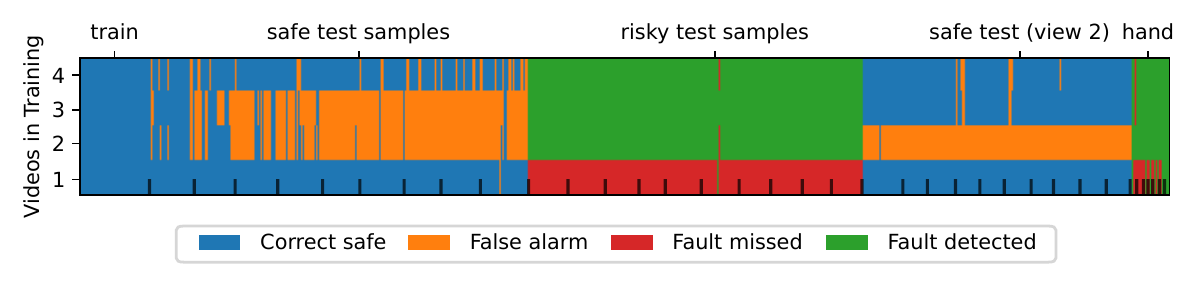}
    \vspace{-2em}
    \caption{Data Aggregation. Generalization of the \textit{Pick Peg} skill as additional labeled videos are incorporated ($y$-axis: number of training videos). Evaluation is performed across 35 test executions ($x$-axis), totaling 8,600 samples. With only one training video, the model fails to detect in-distribution faults but correctly identifies out-of-distribution cases (e.g., ``hands''). Incrementally adding more videos improves generalization across distinct trajectory segments—video 2 helps in detecting risky test samples, while videos 3 and 4 help generalize across the remaining safe test samples.}    
    \label{fig:extra_experiment}
    \vspace{-5pt}
\end{figure}

Finally, Fig. \ref{fig:extra_experiment} illustrates how detection performance scales with additional demonstration videos. When trained on all four available demonstrations, our system generalized effectively across 35 test executions, achieving 92\% overall accuracy and nearly 100\% recall---meaning it reliably flagged every fault occurrence.

\begin{figure}
    \centering
    \includegraphics[width=0.9\linewidth]{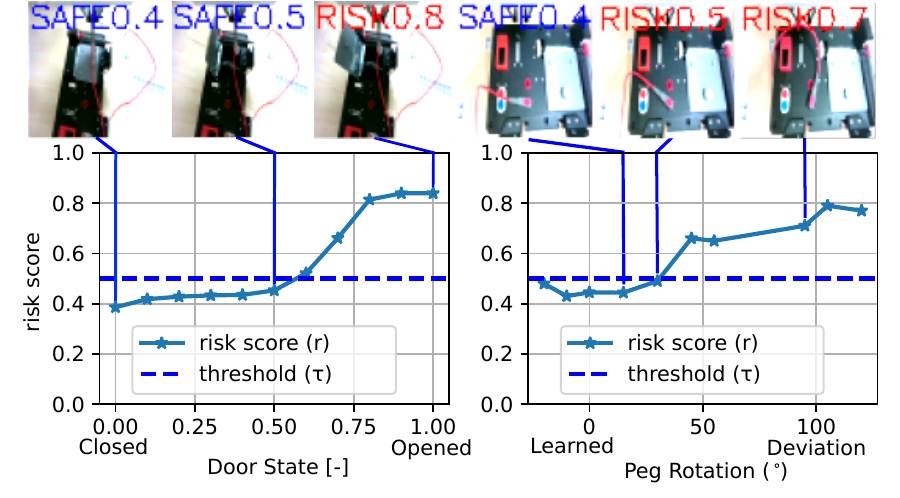}
    \vspace{-1em}
    \caption{(Left) When the door is closed, the situation is safe. As the door begins to open, the risk score increases, reflecting the deviation from the expected state. (Right) The peg can be successfully picked when correctly aligned. As its rotation angle deviates from the trained orientation, the risk score rises. Once this deviation exceeds 30 degrees, the skill fails, triggering the risk flag and stopping the robot.
    }
    \label{fig:oot_exp}
    \vspace{-10pt}
\end{figure}

\textit{Out-of-Distribution Behavior:}
We simulate a transition from a safe to a risky state by manually manipulating the door and peg using an invisible twine to induce unexpected changes. This setup demonstrates how such deviations influence the risk score, as illustrated in Fig.~\ref{fig:oot_exp}.

\begin{figure}[t]
\centering
\hspace{1.5em}
\includegraphics[trim={0cm 0.0cm 0cm 0.0cm},clip,width=0.8\linewidth]{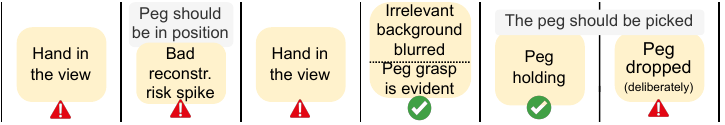}
\includegraphics[trim={1.5cm 0.9cm 0.9cm 1cm},clip,width=0.99\linewidth]{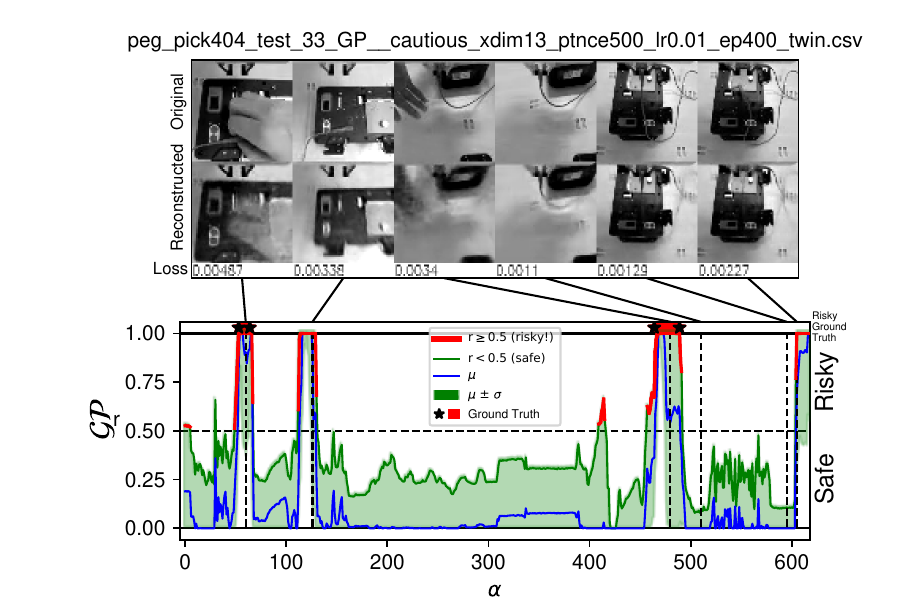}
\includegraphics[trim={1.5cm 0cm 0.9cm 5.45cm},clip,width=0.99\linewidth]{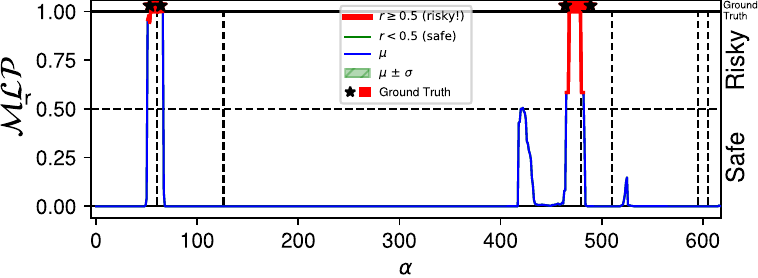}
\vspace{-1em}
\caption{Risk score prediction for a test trial of the \textit{Pick Peg} skill, highlighting a novel (out-of-distribution) fault and the impact of poor video reconstruction. Predictions from the GP model are plotted, with the red curve indicating triggered risk flags ($r > \tau$), and \greenfilledstar{} marking ground-truth risky segments. Below the plot, original and reconstructed frames are displayed along with their reconstruction losses. Note that the second frame from the left exhibits a poor reconstruction, omitting the peg and thus incorrectly triggering a risk flag. In the third frame, a novel fault occurs as a human hand enters the field of view, causing an additional rise in the risk score. At the end, the gripper intentionally remains open. The system employs a non-interrupting policy, allowing execution to continue despite detected risks.}
\label{fig_modelcomparison2}
\end{figure}

\subsection{Impact of Reconstruction Quality}
\label{sec:exp_reconstruction}

The quality of video reconstruction directly affects fault-detection performance. As shown in Fig.~\ref{fig_modelcomparison2} (second frame from the left), low-quality reconstructions frequently lead to misclassifications (false alarms). To mitigate this issue, the autoencoder should be fine-tuned with new images after each run.

We evaluated latent-space dimensions ranging from 8 to 64. Dimensions of 12 or higher enabled the AE to reconstruct fault events over sequences of up to 700 frames. However, smaller dimensions (e.g.,\ 12) often blurred critical details, such as variations in peg rotation, because the AE is trained with an unsupervised reconstruction loss that does not explicitly emphasize these features. Larger latent spaces (48 or 64) captured a broader range of peg rotations but still fell short of fully generalizing across an effectively infinite continuum of angles. In constrained settings with only a few discrete states, such as a door being either open or closed, even a moderate latent dimensions produce sufficiently accurate reconstructions to reliably detect faults.

\subsection{Quality of The Risk Estimation for Novel Risks}
\label{sec_experiment_2}

In this experiment, we assess our model’s ability to detect novel faults, i.e., faults not encountered during training. The dataset (Sec.~\ref{sec:datasets}) includes examples of unexpected objects entering the scene, such as a human hand or tangled cables..

Our GP risk estimator (Sec.\ref{sec:risky_safe}, Eq.\eqref{eq:novel_risk}) flags any significant deviation from the learned skill (safe distribution) as risky. The out-of-distribution threshold is determined automatically via the Automatic Relevance Determination (ARD) mechanism (Eq.~\eqref{eq:ard}), which learns this threshold to separate safe from risky image embeddings based on the training data.

Figure~\ref{fig_modelcomparison2} illustrates the GP model’s response when a human hand enters the scene. As soon as the hand appears, the GP’s predictive uncertainty $\sigma^*$ sharply increases, along with the risk score $r$, accurately signaling the emergence of an unfamiliar scenario.

We found that smaller latent dimensions ($\leq 16$) yield sparser connections among training samples, preserving longer length scales and thus maintaining lower uncertainty for familiar frames. In contrast, latent dimensions larger than 16 collapse length scales too rapidly, causing uniformly high uncertainty, even for images closely resembling training data.

By contrast, an MLP risk estimator (Table~\ref{tab:method_comparison}, ``Novel Faults'' subset) failed to recognize hands or other unexpected objects, posing a significant safety risk in real-world deployments.

\section{Conclusion}
\label{sec:conclusion}

In this paper, we introduced a method to detect faults during robotic manipulation and prevent task failures. We demonstrate our approach in a Learning-from-Demonstration setting, where a robot is taught manipulation tasks with an Electronic Task Board. A human supervisor subsequently observes several executions of the learned skills and labels potentially risky situations. Using a Gaussian Process (GP) model, our method can identify both known and previously unseen faults.

We experimentally compare our method with logistic regression (LR) and an optimized multi-layer perceptron (MLP). While the GP and MLP models perform comparably when detecting previously seen, labeled faults, the GP model shows a clear advantage in handling novel situations. Specifically, scenarios involving previously unseen faults, such as the unexpected appearance of a human hand in the scene, were not consistently recognized by the MLP. In contrast, the GP model effectively detected these out-of-distribution events thanks to its inherent uncertainty estimation capability.

In our future research, we will generalize the method further by incorporating additional modalities into the risk estimation, such as force sensors or microphone input. This enhancement can increase robot's situational awareness and expand the range of tasks in which robots can act autonomously and safely.

\bibliographystyle{IEEEtran}
\bibliography{root}

\begin{thebibliography}{10}
\providecommand{\url}[1]{#1}
\csname url@samestyle\endcsname
\providecommand{\newblock}{\relax}
\providecommand{\bibinfo}[2]{#2}
\providecommand{\BIBentrySTDinterwordspacing}{\spaceskip=0pt\relax}
\providecommand{\BIBentryALTinterwordstretchfactor}{4}
\providecommand{\BIBentryALTinterwordspacing}{\spaceskip=\fontdimen2\font plus
\BIBentryALTinterwordstretchfactor\fontdimen3\font minus \fontdimen4\font\relax}
\providecommand{\BIBforeignlanguage}[2]{{%
\expandafter\ifx\csname l@#1\endcsname\relax
\typeout{** WARNING: IEEEtran.bst: No hyphenation pattern has been}%
\typeout{** loaded for the language `#1'. Using the pattern for}%
\typeout{** the default language instead.}%
\else
\language=\csname l@#1\endcsname
\fi
#2}}
\providecommand{\BIBdecl}{\relax}
\BIBdecl

\bibitem{loborg1994error}
P.~Loborg, ``Error recovery in automation-an overview,'' in \emph{AAAI Symp. on Detecting and Resolving Errors in Manuf. Syst.}, 1994, pp. 94--100.

\bibitem{Ruiz-Celada_Dalmases_Zaplana_Rosell_2024}
O.~Ruiz-Celada, A.~Dalmases, I.~Zaplana, and J.~Rosell, ``Smart perception for situation awareness in robotic manipulation tasks,'' \emph{IEEE Access}, vol.~12, p. 53974–53985, 2024.

\bibitem{DERNER2021103676}
E.~Derner, C.~Gomez, A.~C. Hernandez, R.~Barber, and R.~Babuška, ``Change detection using weighted features for image-based localization,'' \emph{Robotics and Autonomous Systems}, vol. 135, p. 103676, 2021.

\bibitem{Van_Ge_Ren_2017}
M.~Van, S.~S. Ge, and H.~Ren, ``Finite time fault tolerant control for robot manipulators using time delay estimation and continuous nonsingular fast terminal sliding mode control,'' \emph{IEEE Trans. on Cybernetics}, vol.~47, no.~7, p. 1681–1693, Jul. 2017.

\bibitem{Wu_Luo_Zeng_Li_Zheng_2016}
L.~Wu, W.~Luo, Y.~Zeng, F.~Li, and Z.~Zheng, ``Fault detection for underactuated manipulators modeled by markovian jump systems,'' \emph{IEEE Trans. on Ind. Electr.}, vol.~63, no.~7, p. 4387–4399, Jul. 2016.

\bibitem{chandola2009anomaly}
V.~Chandola, A.~Banerjee, and V.~Kumar, ``Anomaly detection: A survey,'' \emph{ACM computing surveys}, vol.~41, no.~3, pp. 1--58, 2009.

\bibitem{7487160}
D.~Park, Z.~Erickson, T.~Bhattacharjee, and C.~C. Kemp, ``Multimodal execution monitoring for anomaly detection during robot manipulation,'' in \emph{IEEE Int. Conf. on Rob. and Aut. (ICRA)}, 2016, pp. 407--414.

\bibitem{6697200}
E.~Di~Lello, M.~Klotzbücher, T.~De~Laet, and H.~Bruyninckx, ``Bayesian time-series models for continuous fault detection and recognition in industrial robotic tasks,'' in \emph{IEEE/RSJ Int. Conf. on Intell. Rob. and Syst. (IROS)}, 2013, pp. 5827--5833.

\bibitem{kappler2015data}
D.~Kappler, P.~Pastor, M.~Kalakrishnan, M.~W{\"u}thrich, and S.~Schaal, ``Data-driven online decision making for autonomous manipulation.'' in \emph{Robotics: science and systems}, vol.~11, 2015.

\bibitem{hullermeier2021aleatoric}
E.~H{\"u}llermeier and W.~Waegeman, ``Aleatoric and epistemic uncertainty in machine learning: An introduction to concepts and methods,'' \emph{Machine learning}, vol. 110, no.~3, pp. 457--506, 2021.

\bibitem{8279425}
D.~Park, Y.~Hoshi, and C.~C. Kemp, ``A multimodal anomaly detector for robot-assisted feeding using an lstm-based variational autoencoder,'' \emph{IEEE Rob. and Aut. Lett.}, vol.~3, no.~3, pp. 1544--1551, 2018.

\bibitem{sinhaAnomaly2024}
\BIBentryALTinterwordspacing
R.~Sinha, A.~Elhafsi, C.~Agia, M.~Foutter, E.~Schmerling, and M.~Pavone, ``Real-time anomaly detection and reactive planning with large language models,'' pp. 1--24, 07 2024, delft, Netherlands, July 15--19, 2024. [Online]. Available: \url{https://www.roboticsproceedings.org/rss20/p114.html}
\BIBentrySTDinterwordspacing

\bibitem{8594169}
A.~Inceoglu, G.~Ince, Y.~Yaslan, and S.~Sariel, ``Failure detection using proprioceptive, auditory and visual modalities,'' in \emph{2018 IEEE/RSJ Int. Conf. on Intell. Rob. and Syst. (IROS)}, 2018, pp. 2491--2496.

\bibitem{10124202}
D.~Altan and S.~Sariel, ``Clue-ai: A convolutional three-stream anomaly identification framework for robot manipulation,'' \emph{IEEE Access}, vol.~11, pp. 48\,347--48\,357, 2023.

\bibitem{9076630}
R.~Perez-Dattari, C.~Celemin, G.~Franzese, J.~Ruiz-del Solar, and J.~Kober, ``Interactive learning of temporal features for control: Shaping policies and state representations from human feedback,'' \emph{IEEE Robotics \& Automation Magazine}, vol.~27, no.~2, pp. 46--54, 2020.

\bibitem{9636710}
G.~Franzese, A.~Mészáros, L.~Peternel, and J.~Kober, ``Ilosa: Interactive learning of stiffness and attractors,'' in \emph{IEEE/RSJ Int. Conf. on Intell. Robots and Syst. (IROS)}, 2021, pp. 7778--7785.

\bibitem{iilsurvey}
C.~Celemin, R.~P\'{e}rez-Dattari, E.~Chisari, G.~Franzese, L.~de~Souza~Rosa, R.~Prakash, Z.~Ajanovi\'{c}, M.~Ferraz, A.~Valada, and J.~Kober, ``Interactive imitation learning in robotics: A survey,'' \emph{Found. Trends Robot}, vol.~10, no. 1–2, p. 1–197, nov 2022.

\bibitem{daggeroriginal}
S.~Ross, G.~Gordon, and D.~Bagnell, ``A reduction of imitation learning and structured prediction to no-regret online learning,'' in \emph{Proceedings of the Fourteenth International Conference on Artificial Intelligence and Statistics}, ser. Proceedings of Machine Learning Research, G.~Gordon, D.~Dunson, and M.~Dudík, Eds., vol.~15.\hskip 1em plus 0.5em minus 0.4em\relax Fort Lauderdale, FL, USA: PMLR, 11--13 Apr 2011, pp. 627--635.

\bibitem{ThriftyDAgger}
R.~Hoque, A.~Balakrishna, E.~R. Novoseller, A.~Wilcox, D.~S. Brown, and K.~Goldberg, ``Thriftydagger: Budget-aware novelty and risk gating for interactive imitation learning,'' in \emph{CoRL}, 2021.

\bibitem{8793698}
M.~Kelly, C.~Sidrane, K.~Driggs-Campbell, and M.~J. Kochenderfer, ``Hg-dagger: Interactive imitation learning with human experts,'' in \emph{2019 Int. Conf. on Rob. and Autom. (ICRA)}, 2019, pp. 8077--8083.

\bibitem{8968287}
K.~Menda, K.~Driggs-Campbell, and M.~J. Kochenderfer, ``Ensembledagger: A bayesian approach to safe imitation learning,'' in \emph{IEEE Int. Conf. on Intell. Rob. and Syst. (IROS)}, 2019, pp. 5041--5048.

\bibitem{FIRE}
T.~Ablett, F.~Marić, and J.~Kelly, ``Fighting failures with fire: Failure identification to reduce expert burden in intervention-based learning,'' \emph{ArXiv}, vol. abs/2007.00245, 2020.

\bibitem{xu2025detectfailuresfailuredata}
\BIBentryALTinterwordspacing
C.~Xu, T.~K. Nguyen, E.~Dixon, C.~Rodriguez, P.~Miller, R.~Lee, P.~Shah, R.~Ambrus, H.~Nishimura, and M.~Itkina, ``Can we detect failures without failure data? uncertainty-aware runtime failure detection for imitation learning policies,'' 2025. [Online]. Available: \url{https://arxiv.org/abs/2503.08558}
\BIBentrySTDinterwordspacing

\bibitem{curtis2024partiallyobservabletaskmotion}
\BIBentryALTinterwordspacing
A.~Curtis, G.~Matheos, N.~Gothoskar, V.~Mansinghka, J.~Tenenbaum, T.~Lozano-Pérez, and L.~P. Kaelbling, ``Partially observable task and motion planning with uncertainty and risk awareness,'' 2024. [Online]. Available: \url{https://arxiv.org/abs/2403.10454}
\BIBentrySTDinterwordspacing

\bibitem{HintonSalakhutdinov2006b}
G.~E. Hinton and R.~R. Salakhutdinov, ``Reducing the dimensionality of data with neural networks,'' \emph{Science}, vol. 313, no. 5786, pp. 504--507, 2006.

\bibitem{Rasmussen2004}
C.~E. Rasmussen, \emph{Gaussian Processes in Machine Learning}.\hskip 1em plus 0.5em minus 0.4em\relax Berlin, Heidelberg: Springer Berlin Heidelberg, 2004, pp. 63--71.

\bibitem{Deisenroth2015}
M.~P. Deisenroth, D.~Fox, and C.~E. Rasmussen, ``Gaussian processes for data-efficient learning in robotics and control,'' \emph{IEEE Trans. on Patt. Anal. and Mach. Intell.}, vol.~37, no.~2, pp. 408--423, 2015.

\bibitem{Williams05}
C.~E. Rasmussen and C.~K.~I. Williams, \emph{Gaussian Processes for Machine Learning (Adaptive Computation and Machine Learning)}.\hskip 1em plus 0.5em minus 0.4em\relax The MIT Press, 2005.

\bibitem{so2024digital}
P.~So, A.~Sarabakha, F.~Wu, U.~Culha, F.~J. Abu-Dakka, and S.~Haddadin, ``Digital robot judge: Building a task-centric performance database of real-world manipulation with electronic task boards,'' \emph{IEEE Robotics \& Automation Magazine}, 2024.

\bibitem{Koonce2021}
B.~Koonce, \emph{ResNet 50}.\hskip 1em plus 0.5em minus 0.4em\relax Berkeley, CA: Apress, 2021, pp. 63--72.

\bibitem{imagenet}
J.~Deng, W.~Dong, R.~Socher, L.-J. Li, K.~Li, and L.~Fei-Fei, ``Imagenet: A large-scale hierarchical image database,'' in \emph{2009 IEEE Conf. on Comp. Vis. and Patt. Rec.}, 2009, pp. 248--255.

\bibitem{kingma2017adammethodstochasticoptimization}
D.~P. Kingma and J.~Ba, ``Adam: A method for stochastic optimization,'' 2017.

\end{thebibliography}


\end{document}